\documentclass[runningheads]{llncs}
\usepackage[T1]{fontenc}
\usepackage[utf8]{inputenc}
\usepackage{textcomp}
\usepackage{booktabs}
\usepackage{graphicx,verbatim}
\usepackage{amsmath}
\usepackage{amssymb}
\usepackage{float} 
\usepackage{graphicx}
\usepackage{subcaption}
\usepackage{multirow}
\usepackage{makecell}
\usepackage{ulem} 
\usepackage{amssymb}
\usepackage{stfloats}
\usepackage{placeins}
\usepackage[percent]{overpic}
\usepackage[table]{xcolor}
\usepackage{url}

\newcommand{\runin}[1]{\par\noindent\textbf{#1}\ }

\newcommand{\ignore}[1]{}

\begin{document}
%

\title{Measuring What VLMs Don't Say: Validation Metrics Hide Clinical Terminology Erasure in Radiology Report Generation}




\titlerunning{Measuring What VLMs Don't Say}
%
%


\author{Aditya Parikh \and
Aasa Feragen \and
Sneha Das \and
Stella Frank}

\authorrunning{Parikh et al.}

\institute{DTU Compute, Technical University of Denmark}

\maketitle              
\begin{abstract}


Reliable deployment of Vision-Language Models (VLMs) in radiology requires validation metrics that go beyond surface-level text similarity to ensure clinical fidelity and demographic fairness.
This paper investigates a critical blind spot in current model evaluation: the use of decoding strategies that lead to high aggregate token-overlap scores despite succumbing to \textbf{template collapse}, in which models generate only repetitive, safe generic text and omit clinical terminology.
Unaddressed, this blind spot can lead to \textbf{metric gaming}, where models that perform well on benchmarks prove clinically uninformative.
Instead, we advocate for lexical diversity measures to check model generations for clinical specificity.
We introduce Clinical Association Displacement (CAD), a vocabulary-level framework that quantifies shifts in demographic-based word associations in generated reports. 
Weighted Association Erasure (WAE) aggregates these shifts to measure the clinical signal loss across demographic groups.
We show that deterministic decoding produces high levels of semantic erasure, while stochastic sampling generates diverse outputs but risks introducing new bias, motivating a fundamental rethink of how ``optimal'' reporting is defined. Code available at~\cite{ourrepo,ourhuggingface}



\end{abstract}

\section{Introduction and Related Work}

Radiology report generation (RRG) aims to automate the conversion of medical images into clinically actionable text, reduce documentation burden and support diagnostic decision-making~\cite{demner2009can,rocha2020evidence}. While recent VLMs have made RRG more viable in clinical workflows~\cite{nam2025multimodal}, in this paper we demonstrate that there is a major gap between scoring well on standard text similarity metrics and providing better clinical reporting or more equitable outcomes across patient populations.


\textbf{Metrics That Enable Gaming.} Contemporary RRG evaluation relies on word-overlap metrics like BLEU~\cite{papineni2002bleu}, ROUGE~\cite{lin2004rouge}, BERTScore~\cite{zhang2019bertscore}, measuring aggregate agreement between generated and reference (ground-truth) reports. Models evaluated on such metrics are vulnerable to \textbf{template collapse}: generating high-frequency ``normal finding'' templates that maximise token overlap despite omitting patient-specific details~\cite{yun2025price}. As normal radiographs dominate datasets, such generic outputs can achieve deceptively high scores with low diagnostic value~\cite{gao2020handling}. Standard metrics \textbf{cannot detect terminology redistribution}, in which group-specific clinical terms are suppressed, over-generated, or reassigned across demographic groups. 
The risk increases with VLMs' documented tendency to under-utilize visual evidence in favor of language priors~\cite{lee2025vlind}, which can lead to more generic outputs reproducing demographic biases.
Current evaluation protocols focus on aggregate accuracy, leaving such biases undetected.

\textbf{Decoding as a Confounding Variable.} The inference strategy (deterministic/stochastic decoding) directly shapes the diversity and specificity of generated reports~\cite{presacan2026comparative}. Deterministic decoding optimises for mode selection, converging to a small set of high-probability templates; stochastic decoding introduces controlled randomness to explore lower-probability tokens. Prior work~\cite{10.1145/3583780.3614961} has motivated stochastic decoding based on marginal gains in BLEU and ROUGE scores. However, selecting decoding strategies solely on average-metric improvements obscures critical fairness trade-offs: while stochasticity may reduce template collapse, it can simultaneously introduce or amplify demographic associations that were absent in reference reports~\cite{presacan2026comparative}. Existing evaluation frameworks provide no mechanism to audit these lexical shifts.

\textbf{The Need for Complementary Metrics.} Recent work has called for evaluation methods that go beyond surface similarity to measure clinical utility, demographic equity, and semantic fidelity~\cite{du2026testing}. Clinician-aligned scoring (e.g., RaTEScore~\cite{zhao2024ratescore}) offers one avenue by incorporating expert judgment, but it does not directly quantify how models redistribute clinical terminology across demographic groups. Fairness metrics in NLP often rely on group-wise F1 gaps or counterfactual perturbation tests. What is missing is a \textbf{vocabulary-level diagnostic} that explicitly tracks whether the demographic association structure of clinical language is preserved, erased, or distorted during generation.

\textbf{Our Contributions.} We identify decoding strategy as a hidden confounder in clinical evaluation for RRG, and thus treat inference configuration as a primary experimental factor: (i)~We propose \textbf{Clinical Association Displacement}, a vocabulary-level framework quantifying shifts in group-associated word usage between human-authored and model-generated reports, identifying erased terminology as well as emergent and preserved bias; (ii)~We introduce \textbf{Weighted Association Erasure}, a summary statistic that measures the global magnitude of association displacement, diagnosing whether a model systematically avoids clinically informative language; (iii)~Using \textbf{our radiology VLM ReX-QWEN} as a testbed, we conduct an inference sensitivity analysis, showing that deterministic decoding inflates scores via \textit{template collapse} (high erasure) while stochastic decoding recovers vocabulary but may increase bias; (iv) our baseline comparisons reveal fairness failures invisible to conventional metrics.


\textbf{Our work establishes} that the choice of decoding strategy and metric jointly confounds clinical evaluation. By auditing both what \textit{models say} and systematically \textit{don't say}, we enable transparent evaluation of whether RRG progress reflects genuine clinical utility or evaluation artifacts.

\begin{figure*}[h!]
  \centering
    \includegraphics[width=0.85\linewidth]{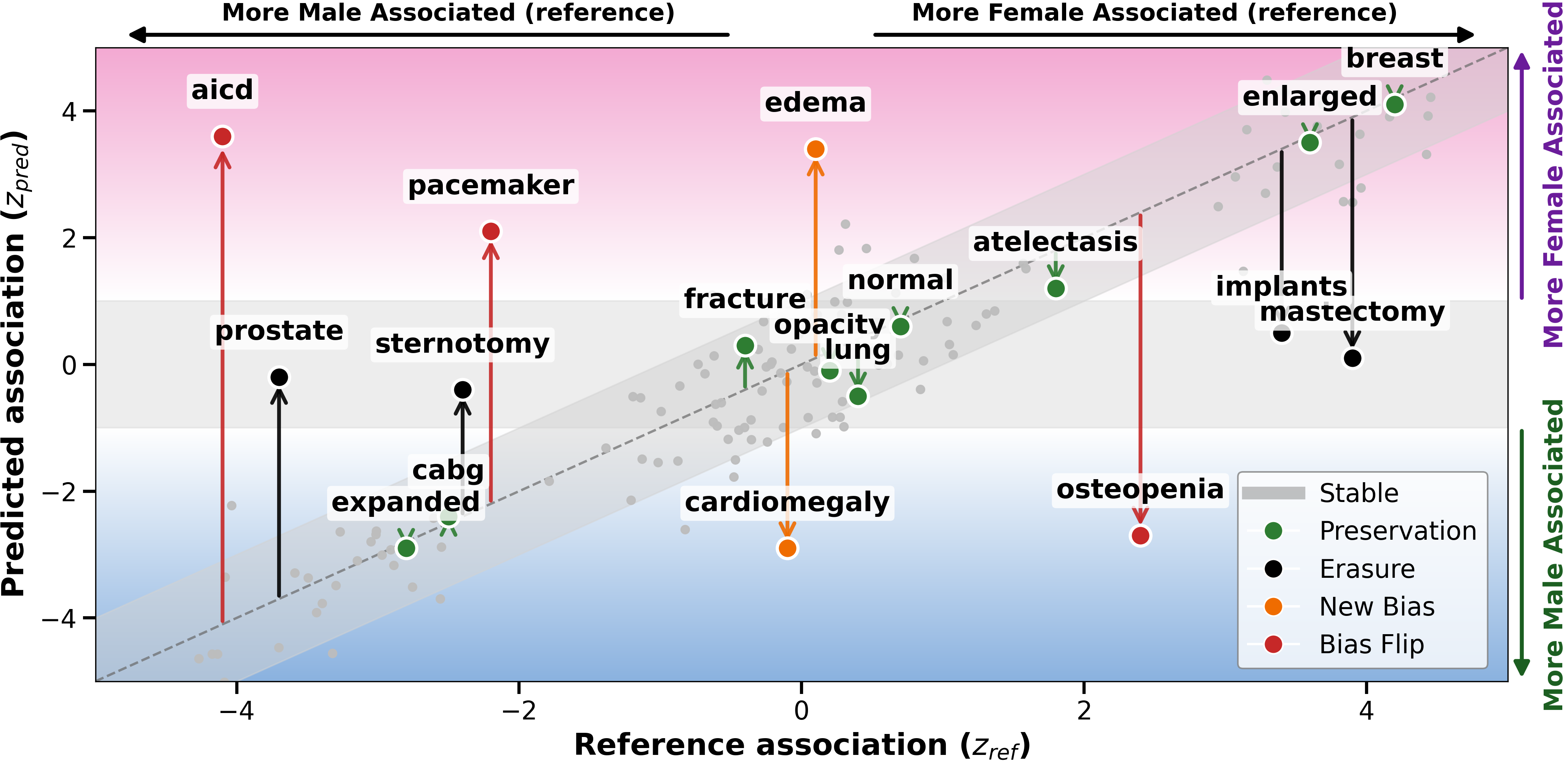}
    \caption{
    \textbf{Clinical interpretation of CAD.} Each point is a vocabulary term, mapped from its demographic association in the reference corpus ($z_{ref}$, x-axis) to its association in predictions ($z_{pred}$, y-axis). Points near the diagonal (gray band) indicate stable associations, while off-diagonal points highlight terms whose demographic association has shifted, reflecting erasure, new bias, or bias flips.
    }
  \label{fig:cad_new}
\end{figure*}

\section{Methods}

We introduce a corpus-level diagnostic of global lexical shift, quantifying how vocabulary is preserved, erased, or newly introduced during generation. Such shifts may involve clinically meaningful terms that are differentially associated with demographic groups and therefore relevant for diagnostic interpretation. 
We therefore also introduce diagnostics for demographic-specific shifts, exposing shifts in lexical usage that are not captured by aggregate metrics.
We formalize this through: Clinical Association Displacement (CAD) (Section~\ref{sec:cad}) and Weighted Association Erasure (WAE) (Section~\ref{sec:wae}).




\subsection{Clinical Association Displacement (CAD)}
\label{sec:cad}

CAD is a vocabulary-level diagnostic framework that quantifies how the demographic association, both in polarity and magnitude, of individual clinical terms shifts between a human-reference corpus and generated outputs (see Fig.~\ref{fig:cad_new}).

\subsubsection{Problem Formulation and Log Ratio:}
Let $\mathcal{D} \in \{\mathrm{ref}, \mathrm{pred}\}$ denote the reference and prediction corpora, where each report is associated with a binary group, here sex: $g \in \{F, M\}$. 
For each corpus $\mathcal{D}$, we compute, for each word $w$ in the vocabulary $\mathcal{V}$, token counts $c_g^{(\mathcal{D})}(w)$ counting the occurrences of $w$ in reports with sex $g$.
The total token count per group is $N_g^{(\mathcal{D})} = \sum_{w \in \mathcal{V}} c_g^{(\mathcal{D})}(w)$.

To measure how strongly a word $w$ is associated with female vs.\ male reports in a corpus $\mathcal{D}$, we use a Dirichlet-smoothed log ratio of unigram probabilities~\cite{monroe2008fightin}.
Let $\alpha>0$ (default: $\alpha$ = 0.1) be a smoothing parameter (where $^\sim$ denotes $\alpha$-smoothed estimates) and $V = |\mathcal{V}|$ the vocabulary size.
We define:

\[
  \forall g\in\{F,M\}:
  \quad p_g^{(\mathcal{D})}(w)=\textstyle \frac{c_g^{(\mathcal{D})}(w)+\alpha}{N_g^{(\mathcal{D})}+\alpha V}
  = \textstyle \frac{\tilde{c}_g^{(\mathcal{D})}(w)} {\sum_{w' \in \mathcal{V}} \tilde{c}_g^{(\mathcal{D})}(w')}, 
  \quad s^{(\mathcal{D})}(w) = \textstyle \log\frac{p_F^{(\mathcal{D})}(w)}{p_M^{(\mathcal{D})}(w)}.
\]

$s^{(\mathcal{D})}(w) > 0$ indicates a \textit{female} association for $w$, while $s^{(\mathcal{D})}(w) < 0$ indicates that $w$ is more associated to \textit{male}.
We apply delta-method approximation under a multinomial model to estimate the variance of $s^{(\mathcal{D})}(w)$:
\[
  \tilde{\mathrm{Var}}[s^{(\mathcal{D})}(w)] \approx \textstyle \frac{1}{\tilde{c}_F^{(\mathcal{D})}(w)} + \frac{1}{\tilde{c}_M^{(\mathcal{D})}(w)},
  \quad
  z^{(\mathcal{D})}(w) = \textstyle\frac{s^{(\mathcal{D})}(w)}{\sqrt{\tilde{\mathrm{Var}}[s^{(\mathcal{D})}(w)]}},
\]
yielding a standardized association $z^{(\mathcal{D})}(w)$ adjusting for frequency variance. We denote $s^{(\mathrm{ref})}(w)$, $s^{(\mathrm{pred})}(w)$ and $z^{(\mathrm{ref})}(w)$, $z^{(\mathrm{pred})}(w)$ for reference and predictions. \\

\noindent\textbf{Quantifying Displacement:}
We define \textbf{Association Displacement $z_{\mathrm{disp}}(w)$} as the standardized difference between prediction and reference associations, and a critical threshold $z_{\mathrm{crit}}$ for significance testing:
\[
  z_{\mathrm{disp}}(w) = \textstyle \frac{s^{(\mathrm{pred})}(w) - s^{(\mathrm{ref})}(w)}{\sqrt{\tilde{\mathrm{Var}}[s^{(\mathrm{pred})}(w)] + \tilde{\mathrm{Var}}[s^{(\mathrm{ref})}(w)]}},
  \quad
  z_{\mathrm{crit}} =\textstyle  \Phi^{-1}\!\left(1 - \frac{p^{\star}}{2}\right).
\]
Treating corpora as independent, $z_{\mathrm{disp}}(w) \sim \mathcal{N}(0,1)$ under the null hypothesis. 
We convert $z_{\mathrm{disp}}(w)$ to two-sided $p$-values, apply Benjamini--Hochberg FDR correction~\cite{benjamini1995controlling}, and flag words with $|z_{\mathrm{disp}}(w)| > z_{\mathrm{crit}}$ at significance level $p^{\star}$ (default: $p^{\star} = 0.05$), where $\Phi^{-1}$ is the standard normal quantile function. 




\subsubsection{Clinical Categories of Shift:} For each vocabulary term we compute $z^{(\mathrm{ref})}$, $z^{(\mathrm{pred})}$ and $z_{\mathrm{disp}}$. Words with $|z_{\mathrm{disp}}|<|z_\mathrm{crit}|$ are deemed \textbf{stable} (no-displacement) and excluded from further audit.
Displaced terms fall into three categories:
\textbf{(i)~Erasure:} A word is strongly associated in the reference ($|z^{(\mathrm{ref})}| > z_{\mathrm{strong}}$) but becomes weak, neutral or is not 
generated ($|z^{(\mathrm{pred})}| < z_{\mathrm{neutral}}$), indicating loss of clinical signal.
\textbf{(ii)~Emergent Bias:} The model introduces group-associations absent in the reference (\textbf{New Bias}: $|z^{(\mathrm{ref})}| < z_{\mathrm{neutral}},\ |z^{(\mathrm{pred})}| \ge z_{\mathrm{neutral}}$) or reverses polarity (\textbf{Bias Flip}: $\mathrm{sign}(z^{(\mathrm{ref})}) \ne \mathrm{sign}(z^{(\mathrm{pred})})$, both non-neutral), indicating hallucinated demographic correlations.
\textbf{(iii)~Preservation:} Polarity is conserved ($\mathrm{sign}(z^{(\mathrm{ref})}) = \mathrm{sign}(z^{(\mathrm{pred})})$) without loss of demographic association. 

We set $z_{\mathrm{neutral}} = 1$ \& $z_{\mathrm{strong}} = 2$ (corresponding to two-tailed normal cutoffs of $p\approx 0.32$ \& $p\approx 0.05$), so only statistically reliable deviations from neutrality are treated as clinically meaningful associations.




\subsection{Weighted Association Erasure (WAE)}
\label{sec:wae}

While CAD tracks displacement for individual words, WAE aggregates these term-level shifts to provide a single, global measure of clinical signal loss across the entire vocabulary, weighted by clinical relevance.

\[
  \mathrm{WAE} = \frac{\sum_{w \in \mathcal{V}} \omega(w) \cdot z_\mathrm{disp}(w)^2}{\sum_{w \in \mathcal{V}} \omega(w)}
\]

\noindent where the numerator represents the \textit{aggregate displacement energy} scaled by term importance (approximated by term frequency). The denominator is a \textit{normalization factor} giving independence of total number of predicted tokens.


We define the weight $\omega(w) = c^{(\mathrm{pred})}(w)$ (prediction-based) or $\omega(w) = c^{(\mathrm{ref})}(w)$ (reference-based). Prediction-based weighting quantifies shifts within text the model \textit{actually produces}, avoiding penalties for omitted vocabulary. Reference-based weighting reveals which clinically important terms (frequent in human reports) are systematically avoided. Aggregating over terms associated with group~$g$ in the reference yields a group-specific score $\mathrm{WAE}_g$, and disparity is defined as $\Delta\mathrm{WAE}$ (e.g., $\Delta\mathrm{WAE} = \Delta\mathrm{WAE}_{F} - \Delta\mathrm{WAE}_{M}$).


Uncertainty is estimated via nonparametric bootstrap; statistical calibration is confirmed via label-permutation tests, where observed WAE consistently exceeds the null distribution under random demographic assignment.


\section{Experiments}

\runin{Dataset: }We use \textbf{ReXGradient-160K}, the largest publicly available multi-site chest X-ray dataset~\cite{zhang2025rexgradient}, which includes clinician-authored findings and impressions. We filter frontal views using a foundational model~\cite{ma2025fully} and stratified subset of $N=2,000$ with balanced sex representation. \\

\runin{Models: ReX-QWEN:} We fine-tune Qwen3-VL-8B-Instruct~\cite{qwen3technicalreport} on ReXGradient using LoRA~\cite{hu2022lora}(rank $r=16$, $\alpha=32$) targeting attention and MLP layers. Input images are resized to $512 \times 512$ RGB. Our primary model, \textbf{ReX-BB} (Balanced Baseline) is trained on 60K sex-balanced examples (50\% male, 50\% female) as our primary model. Two sex-skewed variants: ReX-MB (67\% male) and ReX-FB (67\% female) validate CAD/WAE sensitivity to demographic shifts. We compare ReX-QWEN against three models under identical evaluation protocols: \textbf{CXR-LLaVA}~\cite{lee2025cxr}, \textbf{LLaVA-Rad}~\cite{zambrano2025clinically} (medical VLMs for radiology reporting); and \textbf{OpenAI GPT-4.1}~\cite{openai_gpt41_2025}, prompted zero-shot for report generation. \\

\runin{Inference Strategy Study: } \label{sec:inference_exp} We hypothesise that semantic erasure stems from inference strategies that systematically suppress clinical terminology to minimise generation risk. We partition inference strategies into \textbf{deterministic} (greedy, beam search), which select the highest-probability sequences and may lead to template collapse, and \textbf{stochastic} (temperature $T$, top-$k$, nucleus sampling), which sample from truncated distributions to increase lexical diversity.

Specifically, we evaluate: (i)~Greedy (argmax), (ii)~Beam Search ($k=4$), (iii)~Conservative Sampling ($T=0.3$, top-$p=0.95$), (iv)~Rich Sampling ($T=0.7$, top-$p=0.95$), (v)~Top-$k$ ($k=50$, $T=1.0$), and (vi)~Nucleus Sampling (top-$p=0.9$, $T=1.0$). All use max tokens=512 with caching enabled. \\


\runin{Evaluation Metrics: } \label{sec:eval_metrics} We evaluate models across three dimensions: \textbf{Clinical Accuracy:} We report BERTScore (F1, Precision, Recall) using CheXbert as the backbone model~\cite{smit2020chexbert}, ROUGE-1/L, and RateScore to measure semantic and surface-level similarity against reference reports. \textbf{Textual Diversity:} Following~\cite{guo2024curious}, we compute: (i)~Template Diversity (fraction of unique reports), (ii)~Type-Token Ratio (TTR), (iii)~1-Self-BLEU, and (iv)~Semantic Diversity. \textbf{Our Metrics:} We report: (i)~our proposed CAD-flagged clinical categories (Erasure, New Bias, Bias Flip) quantifying vocabulary-level association shifts, (ii)~our proposed WAE summarizing global displacement magnitude, (iii)~the existing WEAT~\cite{caliskan2017semantics,JMLR:v26:22-1133} and its effect size (ES) as a comparative word-association baseline, and (iv)~Disparity Impact Ratio ($\Delta$DIR)~\cite{2025investigating}, the change in male-to-female token ratio between reference and prediction. Stop words are removed prior to all association-based metric computation to ensure focus on clinical terms.

\section{Results and Discussion}





\begin{figure}[t]
  \centering
  \footnotesize

  \begin{minipage}[b]{0.63\linewidth}
    \centering
    \begin{overpic}[width=\linewidth]{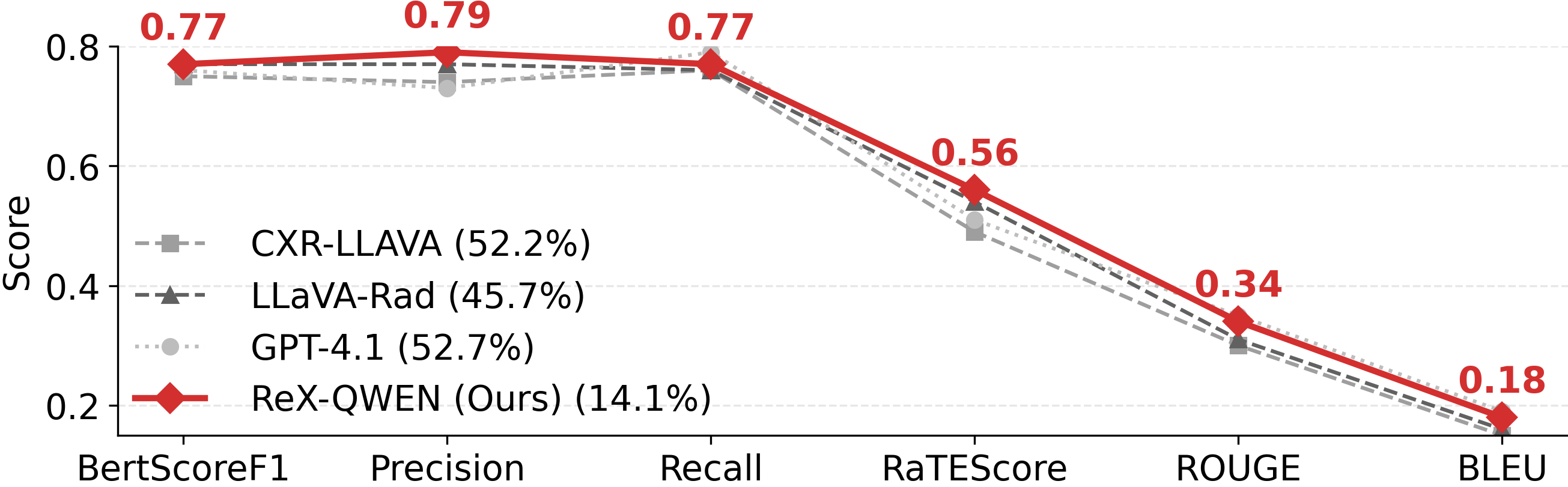}
      \put(7,31){\small\textbf{(a)}}
    \end{overpic}


    \begin{overpic}[width=\linewidth]{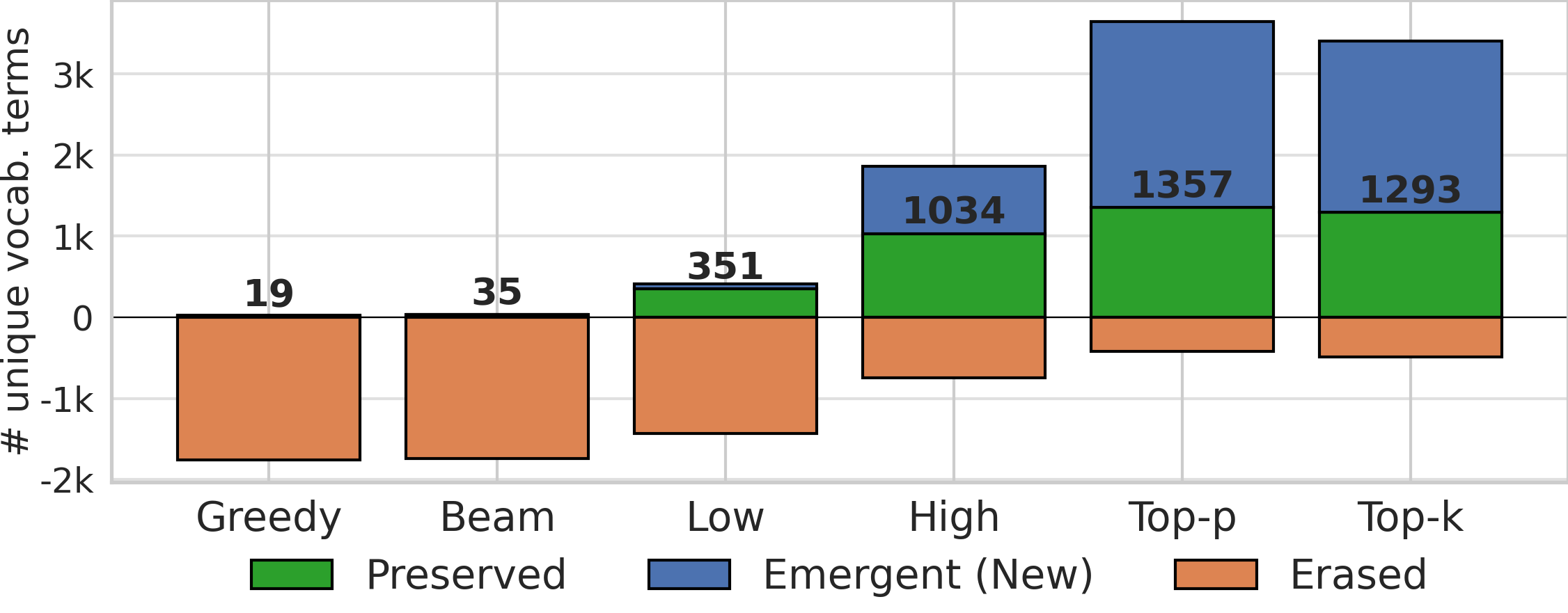}
      \put(7,35){\small\textbf{(b)}}
    \end{overpic}
  \end{minipage}\hfill
  \begin{minipage}[b]{0.36\linewidth}
    \centering
    \begin{overpic}[width=\linewidth]{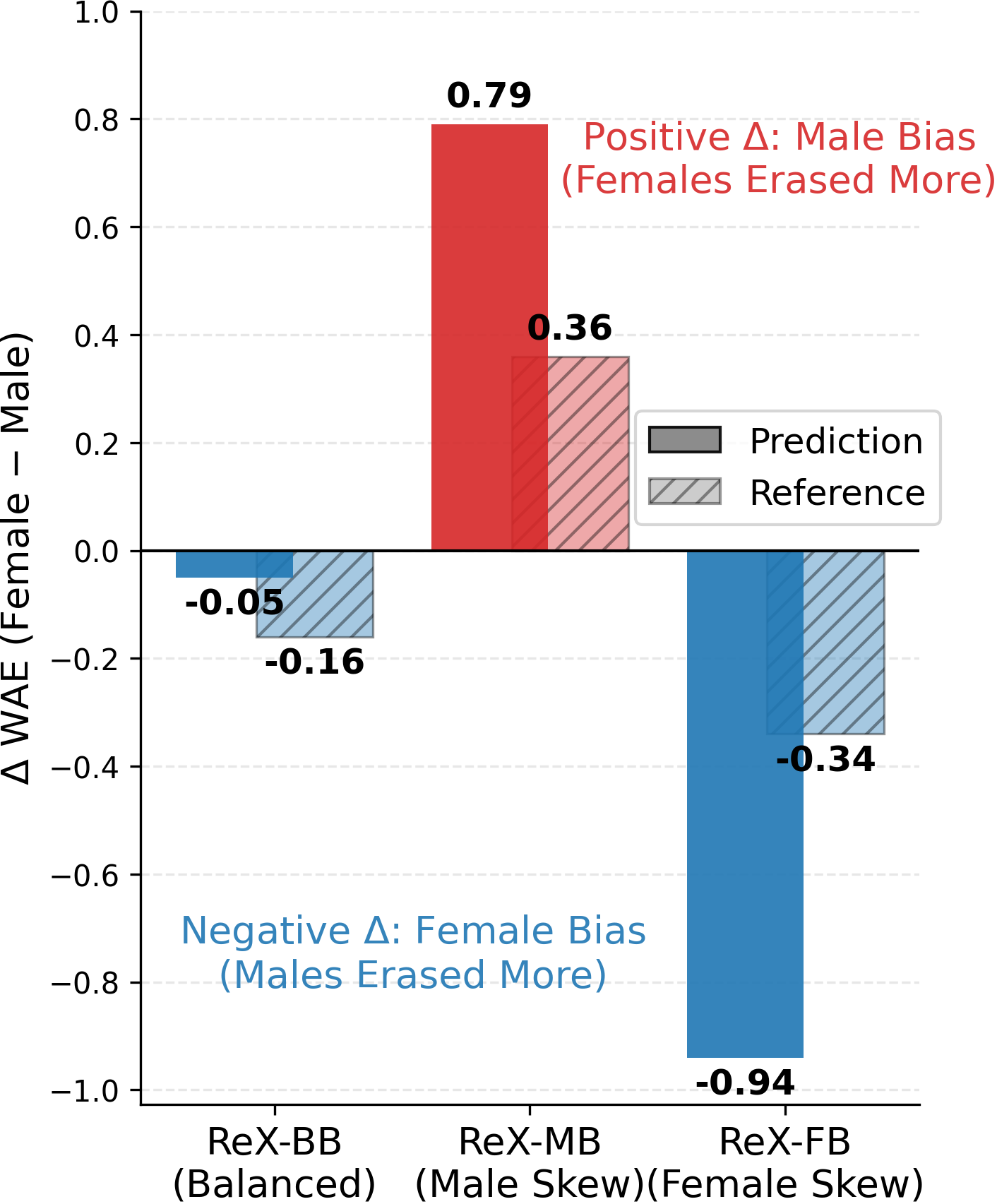}
      \put(10,94){\small\textbf{(c)}}
    \end{overpic}
  \end{minipage}

  \caption{\textbf{(a)}~\textbf{Clinical performance comparison} (w/ Template Diversity \%).
  \textbf{(b)} \textbf{Global lexical shift} across inference strategies.
  \textbf{(c)} $\Delta$WAE successfully captures the semantic shifts caused by sex-skewed training.}
  \label{fig:combined_plots}
\end{figure}

We establish that while ReX-QWEN achieves competitive baseline performance (see Fig.~\ref{fig:combined_plots}a), severe template collapse remains evident (Template Diversity 14.1$-$52.7\%). This collapse motivates our subsequent lexical analysis, specifically investigating how decoding choices drive a fundamental trade-off between semantic erasure and emergent bias. To evaluate these dynamics, we compare ReX-BB across six distinct inference strategies (Table~\ref{tab:inference_sensitivity_partitioned_hline}).

\noindent\textbf{Global lexical collapse.} Deterministic strategies exhibit severe lexical attrition, generating vocabularies of just 22--38 unique terms, 5\% of the original reference vocabulary (see Fig.~\ref{fig:combined_plots}b), collapsing instead to ``normal-finding'' templates. Stochastic approaches preserve vocabulary size but introduce emergent terms absent in references, raising validity concerns: are these recovered clinical signals or hallucinated noise? To determine if these shifts disproportionately affect specific patient groups, we turn to CAD to audit for semantic erasure and emergent bias in sex-associated terms.


\noindent\textbf{Deterministic decoding inflates similarity via collapse.} Greedy and beam search produce 0.3--0.5\% unique reports, presenting a case of \textit{template collapse}. Despite BERTScore F1=$0.78$, CAD reveals severe sex-association erasure: 32--35 erased terms, corresponding to nearly all the highly-biased terms in the references).
For example, \textit{AICD} (male-associated) and \textit{mastectomy} (strongly female-associated) shift their predicted association to neutral. This metric-gaming artifact confirms that high aggregate scores can mask systematic clinical erasure.

\noindent\textbf{Stochastic decoding restores vocabulary but introduces bias.} Temperature scaling increases diversity (36.3--100\%) and reduces erasure but introduces 25--65 new bias terms and elevates WAE$_{pred}$ (2.50 $\rightarrow$ 33.81) due to the emergence of newly biased terms. $\Delta$DIR widens, signaling verbosity imbalances across sex groups. For instance, the pathology \textit{pneumothorax} is neutral in human-reference ($z_{ref} = 0.15$); however low temperature generation actively hallucinated a strong association with male patients ($z_{pred} = -5.69$). WEAT evaluates static latent embeddings and therefore remains entirely blind to decoding-induced biases.

\begin{table}[t]
  \centering
  \caption{\textbf{Inference strategy sensitivity.} (\textbf{Bold}: best; \underline{Underline}: worst).}
  \label{tab:inference_sensitivity_partitioned_hline}
  \setlength{\tabcolsep}{1.0pt}
  \renewcommand{\arraystretch}{1.1}

  \resizebox{\linewidth}{!}{%
    \begin{tabular}{lccccccc}
      \toprule
      \multirow{2}{*}{\textbf{Metric Category}} & \multirow{2}{*}{\textbf{Ref.}} & \multicolumn{2}{c}{\textbf{Deterministic}} & \multicolumn{4}{c}{\textbf{Stochastic (Increasing Entropy)}} \\
      \cmidrule{3-8}
       & & \textbf{Greedy} & \textbf{Beam} & \textbf{$T=0.3$} & \textbf{$T=0.7$} & \textbf{Top-$k$} & \textbf{Top-$p$} \\
      \midrule
      
      \multicolumn{8}{l}{\textit{\textbf{Surface Level Similarity}}} \\
      BERTScore F1 $\uparrow$ & -- & \textbf{0.78} & \textbf{0.78} & 0.76 & 0.73 & 0.71 & \underline{0.69} \\
      BERTScore P/R $\uparrow$ & -- & 0.80/0.75 & \textbf{0.81/0.76} & 0.78/0.75 & 0.73/0.74 & \underline{0.68/0.74} & 0.70/0.73 \\
      ROUGE (1/L) $\uparrow$ & -- & 0.34/0.29 & \textbf{0.35/0.30} & 0.32/0.26 & 0.26/0.19 & \underline{0.20/0.14} & 0.21/0.15 \\
      RateScore $\uparrow$ & -- & \textbf{0.55} & \textbf{0.55} & 0.54 & 0.48 & \underline{0.44} & \underline{0.44} \\
      \midrule

      \multicolumn{8}{l}{\textit{\textbf{Textual Diversity \& Collapse}}} \\
      Template Div. (\%) $\uparrow$ & 82.45 & \underline{0.3} & 0.5 & 36.3 & 97.05 & \textbf{100.0} & \textbf{100.0} \\
      TTR (\%) $\uparrow$ & 2.915 & \underline{0.006} & 0.131 & 0.654 & 2.128 & 2.540 & \textbf{2.781} \\
      1 - Self-BLEU (\%) $\uparrow$ & 35.70 & 1.05 & \underline{0.20} & 10.78 & 33.78 & 46.73 & \textbf{46.84} \\
      Semantic Div. (\%) $\uparrow$ & 85.23 & 12.16 & \underline{5.41} & 55.51 & 83.75 & 85.80 & \textbf{86.19} \\
      \midrule

      \multicolumn{8}{l}{\textit{\textbf{Clinical Categories (Ours)}}} \\
      \cellcolor[gray]{0.9}Erasure (\#) $\downarrow$ & -- & \cellcolor[gray]{0.9}32  & \cellcolor[gray]{0.9}\underline{35}  & \cellcolor[gray]{0.9}17  & \cellcolor[gray]{0.9}4  & \cellcolor[gray]{0.9}5  & \cellcolor[gray]{0.9}\textbf{3}  \\
      \cellcolor[gray]{0.9}New Bias (\#) $\downarrow$ & -- & \cellcolor[gray]{0.9}\textbf{0}  & \cellcolor[gray]{0.9}3  & \cellcolor[gray]{0.9}41  & \cellcolor[gray]{0.9}45  & \cellcolor[gray]{0.9}25  & \cellcolor[gray]{0.9}\underline{65}  \\
      \cellcolor[gray]{0.9}Bias Flip (\#) $\downarrow$ & -- & \cellcolor[gray]{0.9}\textbf{0}  & \cellcolor[gray]{0.9}1  & \cellcolor[gray]{0.9}16 & \cellcolor[gray]{0.9}33  & \cellcolor[gray]{0.9}44  & \cellcolor[gray]{0.9}\underline{69}  \\
      \midrule

      \multicolumn{8}{l}{\textit{\textbf{Fairness}}} \\
      \cellcolor[gray]{0.9}WAE (pred) (Ours) $\downarrow$ & -- & \cellcolor[gray]{0.9}\textbf{2.49} & \cellcolor[gray]{0.9}2.50 & \cellcolor[gray]{0.9}4.95 & \cellcolor[gray]{0.9}33.81 & \cellcolor[gray]{0.9}12.90 & \cellcolor[gray]{0.9}\underline{13.27} \\
      \cellcolor[gray]{0.9}WAE (ref) (Ours) $\downarrow$ & -- & \cellcolor[gray]{0.9}\textbf{0.19} & \cellcolor[gray]{0.9}0.60 & \cellcolor[gray]{0.9}3.43 & \cellcolor[gray]{0.9}9.06 & \cellcolor[gray]{0.9}4.60 & \cellcolor[gray]{0.9}\underline{14.70} \\
      WEAT (S/ES) $\downarrow$ & -- & -1.46/-0.64 & 1.10/0.03 & -1.35/-0.01 & 4.30/-0.07 & -2.13/-0.02 & 0.27/0.00 \\
      $\Delta\mathrm{DIR}$ $\downarrow$ & -- & -0.03 & \textbf{0.01} & -0.08 & -0.01 & -0.20 & \underline{-0.32} \\
      \bottomrule
    \end{tabular}%
  } 
\end{table}

\noindent\textbf{Implications.} CAD identifies two opposing failure modes: \textbf{(i) semantic erasure} under deterministic decoding: systematic vocabulary suppression invisible to BERTScore/ROUGE; \textbf{(ii) emergent bias} under high-temperature sampling: new demographic associations absent in references. No single strategy simultaneously optimises accuracy, diversity, and fairness, motivating fairness-aware decoding selection tuned jointly on clinical accuracy \textit{and} CAD/WAE diagnostics rather than defaulting to greedy inference. \textbf{We recommend} augmenting standard evaluation with CAD categories, WAE, and lexical diversity metrics to prevent interpreting template-collapse artifacts as genuine clinical improvement.

\noindent\textbf{Generalisation:} LLaVA-Rad on ReXGradient and CheXpert~\cite{chambon2024chexpert} confirms the pattern (Table~\ref{tab:llavarad_inference}): deterministic decoding reduces template diversity (45.7\% vs. 91.1\%) with higher erasure (19 terms), while stochastic sampling introduces bias flips (5$\rightarrow$12). WAE drops modestly (1.17$\rightarrow$1.03 on ReXGradient), confirming the trade-off persists across architectures and datasets.

\noindent\textbf{Robustness:} Sex-skewed training (ReX-MB, ReX-FB) leaves conventional metrics stable (BERTScore F1$\in[0.76,0.78]$, RateScore$\in[0.54,0.57]$, WEAT-ES$\approx0$). However, $\Delta\mathrm{WAE}_{\text{pred}}$ goes from near-zero (ReX-BB: -0.05) to $+0.79$ for ReX-MB and $-0.94$ for ReX-FB, confirming WAE captures semantic shifts invisible to existing metrics (refer Fig.~\ref{fig:combined_plots}c). 

\begin{figure}[t]
\centering
  \includegraphics[width=\linewidth]{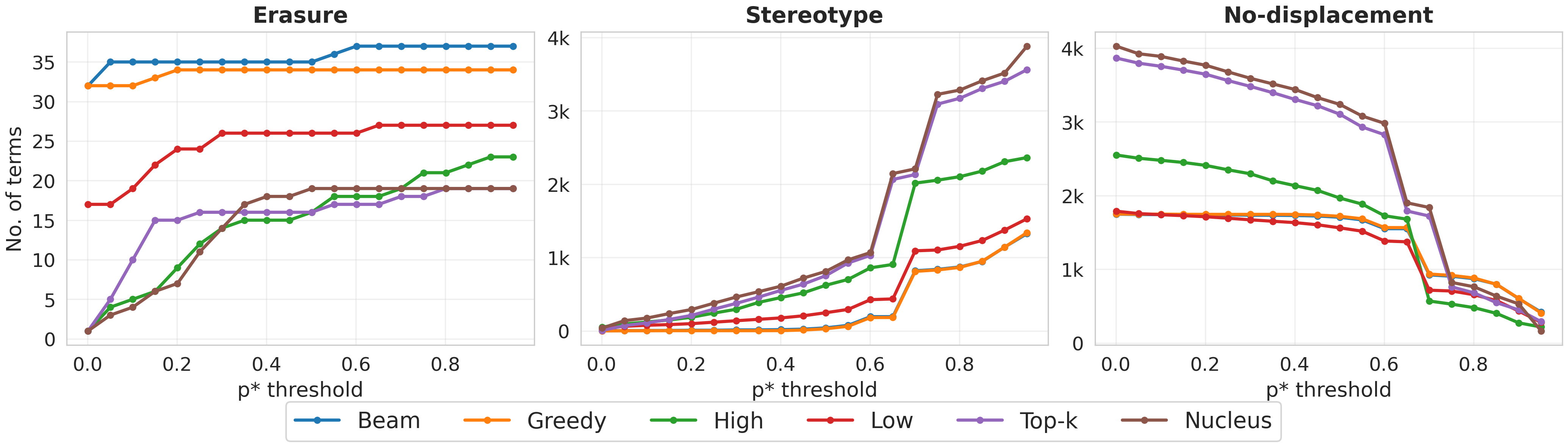}
\caption{\textbf{Ablation}: Sweeping $p_{\mathrm{disp}}^{\star}$ preserves relative model ranking: CAD detects more emergent stereotypes in stochastic outputs across the full range of significance levels.}
\label{fig:ablation_p_value}
\end{figure}

\noindent\textbf{Ablation:} Sweeping $p_{\mathrm{disp}}^{\star}$ across its full range strictly preserves the relative ranking of all decoding strategies (refer Fig.~\ref{fig:ablation_p_value}). This ablation confirms that the erasure-versus-bias trade-off is a fundamental generation mechanic, rather than an artifact of metric thresholding.

\begin{table}[tp]
  \centering
  \footnotesize
  \setlength{\tabcolsep}{2pt} 
  \caption{\textbf{External validation confirms the persistence of the erasure-versus-bias trade-off using LLaVA-Rad with different decoding strategies} Text similarity metrics (BERTScore, ROUGE etc.) are comparable ($\Delta \le 2\%$, unshown), while Diversity metrics show clear differences in template usage (Div. Temp.) and erasure rates. (NB: New Bias; BF: Bias Flip)}
  \label{tab:llavarad_inference}
  
  \resizebox{0.8\linewidth}{!}{%
  \begin{tabular}{lcc|cc}
    \toprule
    \multirow{2}{*}{\textbf{Metrics}}
    & \multicolumn{2}{c}{\textbf{ReXGradient}}
    & \multicolumn{2}{c}{\textbf{CheXpert}} \\
    \cmidrule{2-3}\cmidrule{4-5}
    & \textbf{Greedy} & \textbf{Rich}
    & \textbf{Greedy} & \textbf{Rich} \\
    \midrule


    Div. (Temp./Sem.\%) $\uparrow$ & 45.70/70.32 & 91.05/79.49 & 94.05/80.60 & 100/83.13 \\
    \cellcolor[gray]{0.9}Erasure/NB/BF $\downarrow$ & \cellcolor[gray]{0.9}19/6/5 & \cellcolor[gray]{0.9}11/5/11 & \cellcolor[gray]{0.9}6/1/0 & \cellcolor[gray]{0.9}1/1/4 \\
    \midrule

    \cellcolor[gray]{0.9} WAE (pred/ref) $\downarrow$ & \cellcolor[gray]{0.9}1.17/0.84 & \cellcolor[gray]{0.9}1.03/0.93 & \cellcolor[gray]{0.9}0.76/0.60 & \cellcolor[gray]{0.9}0.84/0.69 \\
    WEAT (S/ES) $\downarrow$ & 1.13/0.01 & -0.36/-0.00 & -0.78/0.11 & -0.14/0.17 \\
    
    \bottomrule
  \end{tabular}%
  }
\end{table}

\section{Conclusion and Future Work}

\noindent\textbf{Future Work: }Our analysis focuses on binary sex associations; future work should extend CAD to age, ethnicity, and intersectional groups. Clinician validation is needed to confirm whether CAD-flagged erasures correspond to clinically meaningful omissions.

\noindent\textbf{Conclusion: }We demonstrate that decoding strategy and metric choice jointly confound evaluation in radiology report generation. By introducing CAD and WAE, we provide a framework to audit semantic erasure and emergent bias that remain completely invisible to conventional metrics. Together, these tools enable transparent evaluation, allowing the field to distinguish genuine clinical utility from the illusion of performance caused by \textit{template collapse}.




%
%
%

\bibliographystyle{splncs04}
\bibliography{bibliography}

\end{document}